\documentclass[letterpaper, 10 pt, conference]{ieeeconf}  

\IEEEoverridecommandlockouts                              

\usepackage{multirow}
\usepackage{graphicx}
\usepackage{amsmath} 
\usepackage{algorithm}
\usepackage{algorithmic}
\usepackage{cite}
\usepackage{color}
\usepackage{flushend}

\title{\LARGE \bf
Semantics-Guided Moving Object Segmentation with 3D LiDAR
}

\author{Shuo Gu$^{1}$, Suling Yao$^{1}$, Jian Yang$^{1}$ and Hui Kong$^{2}$
\thanks{$^1$The authors are with School of Computer Science and Engineering, Nanjing University of Science and Technology, Nanjing, China 210094. Email:{\tt\small \{shuogu,sulingyao,csjyang\}@njust.edu.cn}}
\thanks{$^2$Hui Kong is with Department of Electromechanical Engineering, The University of Macau, Macau, China. Email:{ \tt\small huikong@um.edu.mo}}
}

\begin{document}

\maketitle
\thispagestyle{empty}
\pagestyle{empty}

\begin{abstract}
Moving object segmentation (MOS) is a task to distinguish moving objects, e.g., moving vehicles and pedestrians, from the surrounding static environment. The segmentation accuracy of MOS can have an influence on odometry, map construction, and planning tasks. In this paper, we propose a semantics-guided convolutional neural network for moving object segmentation. The network takes sequential LiDAR range images as inputs. Instead of segmenting the moving objects directly, the network conducts single-scan-based semantic segmentation and multiple-scan-based moving object segmentation in turn. The semantic segmentation module provides semantic priors for the MOS module, where we propose an adjacent scan association (ASA) module to convert the semantic features of adjacent scans into the same coordinate system to fully exploit the cross-scan semantic features. Finally, by analyzing the difference between the transformed features, reliable MOS result can be obtained quickly. Experimental results on the SemanticKITTI MOS dataset proves the effectiveness of our work.  
\end{abstract}

\section{Introduction}
Distinguishing between moving and static parts in the surrounding environment is an important task for autonomous vehicles. Compared with static objects that occupy most of the space, moving objects usually account for only a small part. This unbalance distribution increases the difficulty of the MOS task. In addition, considering the influence of the MOS accuracy on navigation safety, pose estimation, mapping, and path planning tasks, reliable and real-time moving object segmentation is actually a necessary and challenging task. 

\begin{figure}[!t]
\begin{center}
\makebox[1pt]{\includegraphics[width=\linewidth]{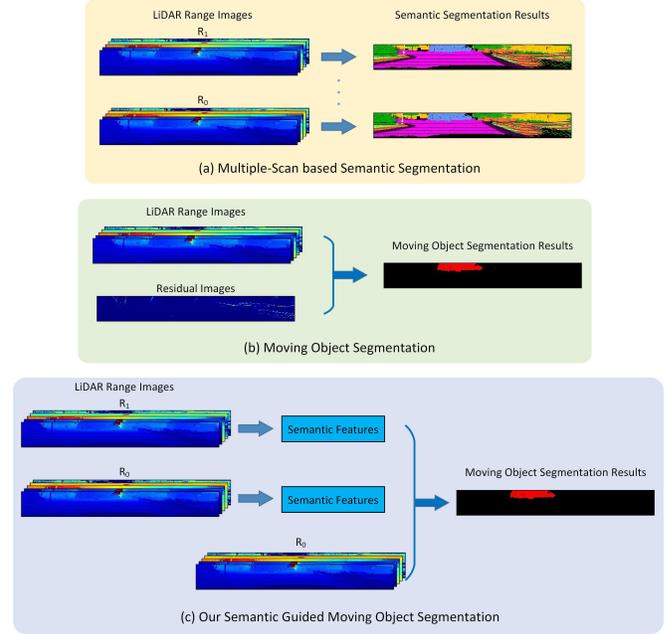}}
\end{center}
\caption{Range-based LiDAR segmentation methods. (a) Multiple-scan-based LiDAR semantic segmentation method. It uses consecutive range images as inputs and outputs corresponding semantic segmentation results, including moving information. (b) LiDAR-based moving object segmentation method. It takes range images and residual images as inputs and outputs the MOS results. (c) Our semantics-guided moving object segmentation method. It includes a single-scan-based semantic segmentation module and a multiple-scan-based MOS module. The latter takes range images and semantic features of the former as inputs. Note that to save space, only the middle size of 64 $\times$ 512 is displayed instead of the full size. This similarly applies to the rest figures of the paper.}
\label{fig:flowchart}
\end{figure}

The existing MOS methods are mainly based on camera images. The LiDAR sensors are rarely used due to the lack of labeled MOS data. In recent years, with the release of the SemanticKITTI MOS dataset, the LiDAR-based MOS methods have gradually attracted more and more attention. In this paper, we focus on developing a MOS network with LiDAR data only. The MOS task can be regarded as a simplified version of the multiple-scan-based semantic segmentation one. The latter not only predicts the semantic classes of each LiDAR point but also determines whether it is moving or not. However, in practical applications, the semantic classes of the moving objects, such as vehicles and pedestrians, have no significant difference in the decision-making process of autonomous vehicles. Therefore, this paper only focuses on the separation of moving and static objects. 

The LiDAR-based MOS methods usually use range images and residual images as inputs and conduct the MOS task directly. By analyzing the multiple-scan-based semantic segmentation methods, we find that the semantic cues are more useful for the MOS task than the commonly used raw LiDAR data. Therefore, instead of dealing with the MOS task alone, we associate the MOS task with the single-scan-based semantic segmentation task. This is similar to the multiple-scan-based semantic segmentation methods. The unified method can predict not only the moving labels but also the semantic ones. Compared with the multiple-scan-based semantic segmentation methods, the combination of the MOS and single-scan-based semantic segmentation has the following advantages: First, the complex multiple-scan-based semantic segmentation task is divided into modules of commonly researched single-scan-based semantic segmentation and relatively simple moving object segmentation. This can make full use of the existing single-scan-based semantic segmentation network architectures. Second, the modular design allows us to directly use the pre-trained single-scan-based models, instead of training from scratch. This simplifies the training process, we can only train the MOS module. The single-scan-based semantic segmentation methods can also upgrade the MOS capability in a quick time without much modification. 

In addition, in order to effectively connect the single-scan-based semantic segmentation module and the MOS module, we propose a novel adjacent scan association (ASA) module to explore the semantic feature association of adjacent LiDAR scans. The ASA module can establish correspondences of the semantic features according to the relative poses between scans. This helps to accurately transfer information between scans. The semantic features are transformed into the same coordinate system and the semantic differences can be used in the MOS task.

In sum, we propose a semantics-guided CNN model for the MOS task. The network consists of three modules: a single-scan-based semantic segmentation module, an adjacent scan association (ASA) module, and a moving object segmentation (MOS) module. As a key component, the ASA module associates the semantic features from the semantic segmentation module, then transforms and transmits them to the MOS module. 

The main contributions of this paper are: (i) The semantic aware MOS simplifies the training process by making use of the existing single-scan-based semantic segmentation network architectures and pre-trained models. (ii) The ASA module realizes accurate association between semantic features of different LiDAR scans. (iii) The modular design enables the upgrading of a single module.
Experimental results on the SemanticKITTI MOS dataset show that the proposed network\footnote{https://competitions.codalab.org/competitions/28894\#results} can obtain accurate MOS performance in a simple and fast way.

\section{Related Works}

\begin{figure*}[!t]
\begin{center}
\makebox[1pt]{\includegraphics[width=\linewidth]{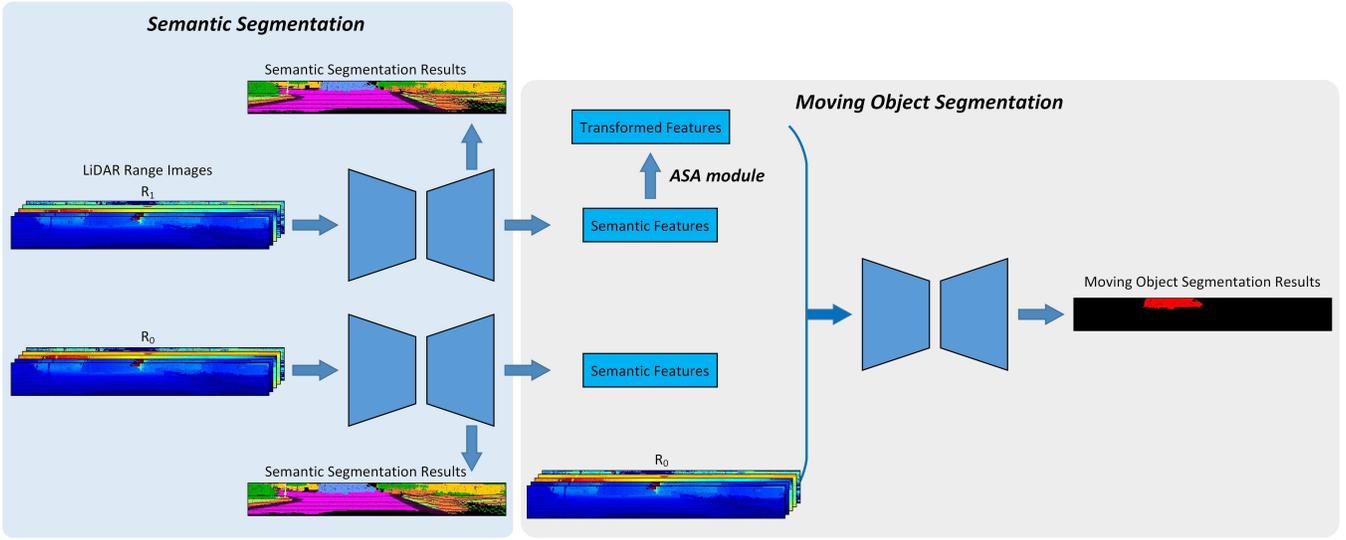}}
\end{center}
\caption{The flowchart of the proposed semantics-guided moving object segmentation method from LiDAR point cloud. The whole network consists of three modules and conducts single-scan-based semantic segmentation and multiple-scan-based moving object segmentation in a cascaded way. The semantic segmentation module learns the semantic features of each LiDAR point. The adjacent scan association module converts the features of different scans to the same coordinate system. The moving object segmentation module combines the transformed semantic features and LiDAR range images to differentiate the moving objects. }
\label{fig:flowchart2}
\end{figure*}

\subsection{LiDAR Data Representation}
\label{sec:representation}
LiDAR is an important sensor for autonomous vehicles. The typical LiDAR data representations can be roughly divided into three categories, including point-based, voxel-based, and projection-based techniques.

The point-based methods process the unstructured LiDAR point clouds directly. PointNet\cite{QiSMG17} is the first point-based method that aggregates information through a permutation invariant operation on the raw LiDAR point clouds. Recent works focus on developing special convolution operations and kernels for point clouds, such as PointConv\cite{WuQL19}, TangentConv\cite{TatarchenkoPKZ18} and KPConv\cite{ThomasQDMGG19}. However, due to memory requirements and time complexity, most point-based methods still struggle on large-scale point clouds.

The voxel-based methods transform the LiDAR point clouds into 3D volumetric grids. MinkowskiNet\cite{ChoyGS19} utilizes the sparse 3D convolutions to efficiently process the voxelized LiDAR data. SegCloud\cite{TchapmiCAGS17} applies a 3D Fully Convolutional Neural Network (FCNN) on the voxelized point clouds. Cylinder3D\cite{Zhu0WHM00L21} splits the raw point clouds into cylindrical grids. However, the 3D convolutions are computationally intensive and the voxels are usually sparse, which limits the resolution of voxels and the overall performance.
The projection-based methods project the LiDAR point clouds onto regular 2D images and use the well-researched 2D CNNs. In SqueezeSeg\cite{WuWYK18}, SqueezeSegV2\cite{WuZZYK19} and RangeNet++\cite{MiliotoVBS19}, the spherical projection is performed on the LiDAR data. The point clouds can also be projected into cartesian bird’s eye view (BEV) (VolMap\cite{abs-1906-11873}) or polar BEV (PolarNet\cite{0035ZDYXGF20}). The projection-based methods can achieve real-time performance, but the accuracy is often limited due to the information loss in the projection process.
Considering the computation and time cost, in this work, we use spherical projection to convert the LiDAR data into 2D range images.

\subsection{LiDAR Semantic Segmentation}
Semantic segmentation and MOS are closely relevant tasks. According to the number of scans used, the LiDAR-based semantic segmentation methods can be divided into single-scan-based semantic segmentation and multiple-scan-based ones. 

The single-scan-based semantic segmentation methods\cite{QiYSG17,LandrieuS18,Hu0XRGWTM20,ChengRTLL21} deal with the static LiDAR data and can only find the movable objects, such as vehicles and humans, not the real moving objects. 
The multiple-scan-based semantic segmentation methods\cite{ShiLWH020,DuerrPWB20,CaoLPLYL20,TangLZLLWH20} takes the continuous sequence of LiDAR scans as inputs. They fuse the spatial information of a single scan and the temporal information between adjacent scans to predict the semantic and moving characteristics of each LiDAR point. The multiple-scan-based semantic segmentation methods can be viewed as the combination of the single-scan-based semantic segmentation and the multiple-scan-based MOS. 
Inspired by the multiple-scan-based semantic segmentation methods, we use the off-the-shelf single-scan-based LiDAR semantic segmentation models to assist the MOS process.

\subsection{LiDAR Moving Object Segmentation}
A variety of approaches have been proposed for MOS based on camera images only\cite{BarnesMPP18,PatilBDM20} or with both camera and LiDAR data\cite{YanCMSM14,PosticaRM16}. Recently, the LiDAR-based MOS methods have achieved more and more attention.

Some LiDAR-based MOS methods\cite{KimK20a,PagadANRKY20,SchauerN18} focus on classifying dynamic points by checking the inconsistency between the query scan and the pre-constructed map. However, the static map can only be built in an offline way, which affects the actual deployment of these methods.
In recent years, the map-free MOS methods using only LiDAR data have achieved success. Ruchti et al.\cite{RuchtiB18} predict the point-wise probabilities of points belonging to moving objects by a learning-based method. Chen et al.\cite{ChenLMWGBS21} combines the LiDAR range images and the residual images to exploit the temporal information and differentiate the moving objects. He et al.\cite{HeERR22} proposes a sequential scene flow estimation method to learn the motion information of the point clouds.
In this work, we propose a semantic-aware MOS model. The MOS task is reformulated as cascaded single-scan-based semantic segmentation and multiple-scan-based MOS task. The semantic features will greatly improve the MOS performance.

\section{Method}
The proposed network conducts a single-scan-based semantic segmentation and a multiple-scan-based MOS in turn as shown in Fig. \ref{fig:flowchart2}.

\subsection{Semantic Segmentation}
\label{sec:semantic}
To achieve real-time performance, the single-scan-based semantic segmentation module uses LiDAR range images as inputs.
The LiDAR points are projected onto a 2D image plane according to the yaw and pitch angles. Assuming the LiDAR point is $P=(x,y,z)$, the transformation is defined as follows,
\begin{equation}
\begin{cases}
\theta_{yaw} = arctan(y,x)\\
\theta_{pitch} = arcsin(z/{\sqrt{x^2+y^2+z^2}}) \\
u = 0.5*[(1-\theta_{yaw}/\pi)] \cdot W\\
v = [1-(\theta_{pitch}-f_{down})/f] \cdot H \\
\end{cases}
\label{equ:spherical}
\end{equation}
where $p=(u,v)$ are the corresponding coordinates in the LiDAR range image, $\theta_{yaw}$ and $\theta_{pitch}$ denoting the yaw and pitch angles, $W$ and $H$ representing the width and height of the range image, respectively. $f=f_{up}-f_{down}$ is the vertical field-of-view of the LiDAR sensor.

\begin{figure}[!t]
\begin{center}
\makebox[1pt]{\includegraphics[width=\linewidth]{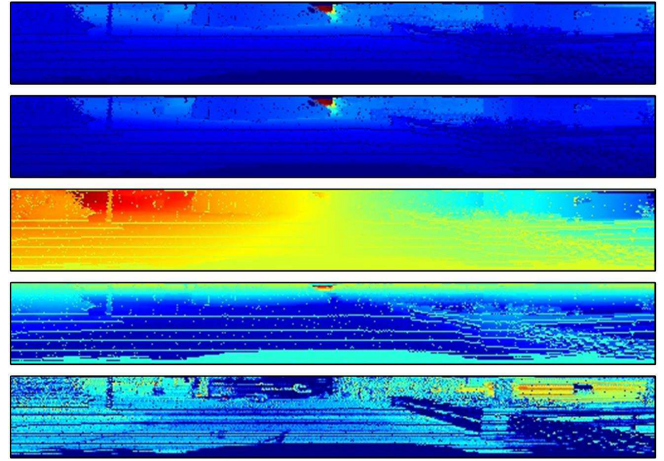}}
\end{center}
\caption{The LiDAR range images include $range$, $x$, $y$, $z$ and $intensity$ components. }
\label{fig:range}
\end{figure}

Based on the correspondences between $P$ and $p$ in Eq.\ref{equ:spherical}, we can get the LiDAR range images of size $H\times W\times C$. $C$ denotes the channel of features. For the Velodyne HDL-64E LiDAR, the size is $64\times 2048\times 5$. The features include $range$ ($\sqrt{x^2+y^2+z^2}$), $x$, $y$, $z$ and the $intensity$. Figure \ref{fig:range} shows examples of the LiDAR range images. 

After the pre-processing step, the LiDAR range images are fed to the single-scan-based semantic segmentation module. To make full use of the existing semantic segmentation models, we directly test our method with the pre-trained models of RangeNet++\cite{MiliotoVBS19} and SalsaNext\cite{CortinhalTA20} in this paper.

\begin{figure}[!t]
\begin{center}
\makebox[1pt]{\includegraphics[width=\linewidth]{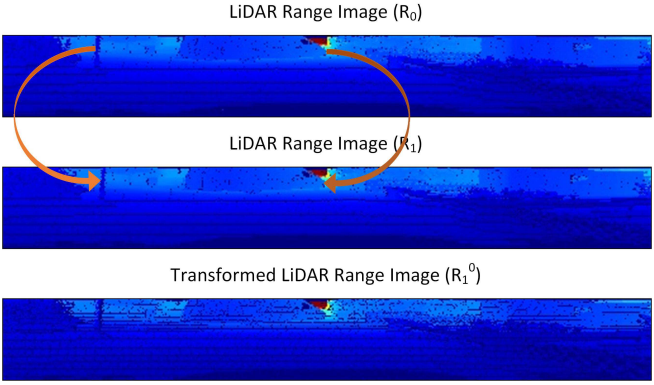}}
\end{center}
\caption{The adjacent scan association module. The arrows denote the correspondences between adjacent LiDAR range images. From top to bottom are the current range image, previous range image, and the transformed range image. }
\label{fig:asa}
\end{figure}

\subsection{Adjacent Scan Association}
After the single-scan-based semantic segmentation module, we can get high-level features of each LiDAR scan. In order to exploit the semantic features of each scan, an intuitive idea may be to concatenate them directly. However, this simple solution does not work because the semantic features are represented in different coordinate systems. In this paper, we propose an adjacent scan association module to convert the semantic features of adjacent scans into the same coordinate system and then feed them to the MOS module. The ASA module consists of three steps: coordinate system conversion, range image generation, and feature transformation.

\textit{Coordinate System Conversion:} 
Assuming that we have a sequence of $N$ LiDAR scans, and the corresponding odometry information is known. $S_0$ denotes the current LiDAR scan and $S_i$ represents one previous scan $(0<i<=N)$. $T_i^0$ denotes the relative transformation between the previous scan $S_i$ and the current scan $S_0$. Given this information, we can convert a LiDAR point in a previous scan $S_i$ to the one in the current scan $S_0$ as follows,

\begin{equation}
\left[ \begin{array}{c} 
x_0^{i} \\ y_0^{i} \\ z_0^{i} \\ 1 \\
\end{array} \right]
=T_{i}^{0} \cdot 
\left[ \begin{array}{c} 
x_i \\ y_i \\ z_i \\ 1 \\
\end{array} \right]
\label{equ:coordinate}
\end{equation}

where $(x_0^{i}, y_0^{i}, z_0^{i}, 1)$ and $(x_i, y_i, z_i, 1)$ denote the coordinates of the same 3D point in the coordinate systems of scan $S_0$ and scan $S_i$, respectively. Considering the semantic features $F_i$ are represented in the form of range image, we only need to transform the LiDAR points lying in range image $R_i$ to the coordinate system of scan $S_0$. 

\textit{Range Image Generation:} 
After the coordinate system conversion step, we can get the transformed LiDAR point clouds. In this step, we use Eq. \ref{equ:spherical} to project the transformed scan $S_0^{i}$ onto the range image $R_0$ and generate a new range image $R_0^{i}$.

Through this $R_i\Rightarrow S_0^{i}\Rightarrow R_0^{i}$ process, we can get a corresponding point $(u_0,v_0)$ in current range image $R_0$ if it exists. The association between the previous and current scans is beneficial to the subsequent MOS task. If the relative pose $T_{i}^{0}$ and the semantic segmentation results are accurate enough, we can even distinguish the moving points from the static by comparing the corresponding semantic segmentation results. The point is static if the semantic prediction in the current scan is the same as that in the previous scan, otherwise, it is moving. However, in practical applications, there exist errors in both relative pose and semantic segmentation results. That is why we use the semantic features rather than the semantic results in this paper.

\textit{Feature Transformation:}
In this step, we use the calculated point correspondence information to assist the feature transformation process. Assuming the association between the previous range image $R_i$ and the current range image $R_0$ is represented by $Tr$ with the size of $H\times W$, $(u_0,v_0)$ is the corresponding point of $(u_i,v_i)$ in $R_0$. The transformation $Tr$ is defined as follows,

\begin{equation}
Tr(u_i,v_i)=
\begin{cases}
u_0+v_0\cdot W\\
0, \quad\text{no correspondences}
\end{cases}
\end{equation} 

where $Tr$ stores the index information from range image $R_i$ to range image $R_0$.

Then, we use the $\mathit{reshape}$ and $\mathit{scatter}$ functions to transform the semantic features $F_i$ of the previous range images $R_i$ to the current range image $R_0$ according to the association information in $Tr$.

\subsection{Moving Object Segmentation}
Like the semantic segmentation module, we reuse the existing single-scan-based semantic segmentation models, RangeNet++\cite{MiliotoVBS19} and SalsaNext \cite{CortinhalTA20}, for the MOS module in this paper. Both RangeNet++ and SalsaNext follow the encoder-decoder architectures and can obtain accurate semantic segmentation results in real-time. After the ASA module, we can get the semantic features of the current range image and the transformed features of the previous range image. As mentioned earlier, the differences between these two semantic features already contain enough information about the movement. However, if only the semantic features are used as inputs, the accuracy of relative poses and semantic segmentation network will limit the MOS performance. Therefore, we also add the range images of the current scan to minimize the negative impact. The MOS module combines the semantic features and range images to differentiate the moving objects. Compared with the range-image-based MOS methods, the addition of transformed semantic features can bring significant improvement.

After the MOS process, we can get the segmentation results in the form of a range image as shown in Fig. \ref{fig:flowchart2}. In order to further improve the performance, the results will be back-projected to the point cloud form. Then, a k-Nearest-Neighbor (kNN) search algorithm is applied to remove the artifacts caused by spherical projection. 

\section{Experiments}

\subsection{SemanticKITTI MOS Dataset}
The SemanticKITTI MOS dataset is a large-scale 3D LiDAR-based moving object segmentation benchmark for the MOS task in outdoor driving scenes. It is built upon KITTI\cite{GeigerLU12} and SemanticKITTI\cite{BehleyGMQBSG19} datasets. The MOS dataset has 22 sequences of LiDAR scans, where sequences 00-07 and 09-10 are used for training, sequence 08 for validation, and sequences 11-21 are used for testing. Only two classes, moving and static, are used in this dataset. The intersection-over-union (IoU) \cite{EveringhamGWWZ10} value on the moving objects is the primary metric used for comparison with other methods.

\subsection{Experimental Setup}
The proposed network is implemented on a server with 64GB RAM, an Intel(R) Xeon(R) E5-2650 CPU, and 2 NVIDIA Geforce RTX2080Ti GPUs under Ubuntu using PyTorch. The evaluation results on the SemanticKITTI MOS testing dataset are obtained based on all training data, with an epoch size of 150 and batch size of 8.
The initial learning rate and the learning rate policy are set to be the same as the original networks (RangeNet++ and SalsaNext). If there is no special description, it indicates that both the single-scan-based semantic segmentation module and the MOS module use the SalsaNext network. The odometry information of each sequence is estimated with the LiDAR-based SLAM method SuMa\cite{BehleyS18}.

\subsection{Different Transformation Designs}
There are two kinds of data representation conversions in the network. One is the transformation of LiDAR scans to range images in the semantic segmentation module, and the other is the transformation of previous semantic features into the current range image in the ASA module. In this section, we will analyze the influences of data type, source data, and hardware on the process of data representation conversion.

\subsubsection{Data Type}
The coordinates of 3D LiDAR points and the relative pose matrices are saved as floating-point numbers, while the coordinates of range images are integer point numbers. This means that the range images in the current coordinate system generated based on previous scans are different from the range images generated based on the current scan. Even if the surrounding scene is static, the relative poses are accurate and there is no occlusion.

\subsubsection{Source Data}
The raw LiDAR scans and the corresponding range images have a different number of points. Therefore, the range images converted from previous scans and previous range images are different.

\subsubsection{Hardware}
Converting 3D LiDAR point clouds into 2D range images inevitably encounters the many-to-one problem, which means multiple LiDAR points may correspond to the same point in the range image. In the semantic segmentation module, we only save the points with minimum range values and perform the transformation on CPUs. However, in the ASA module, the feature transformation occurs on GPUs. Parallel computing cannot ensure that points with minimum range values are saved, and only random points among the multiple points are saved. This characteristic leads to differences between the range images generated based on CPUs and GPUs.

\subsubsection{Analysis} 
\begin{figure}[!t]
\begin{center}
\makebox[1pt]{\includegraphics[width=\linewidth]{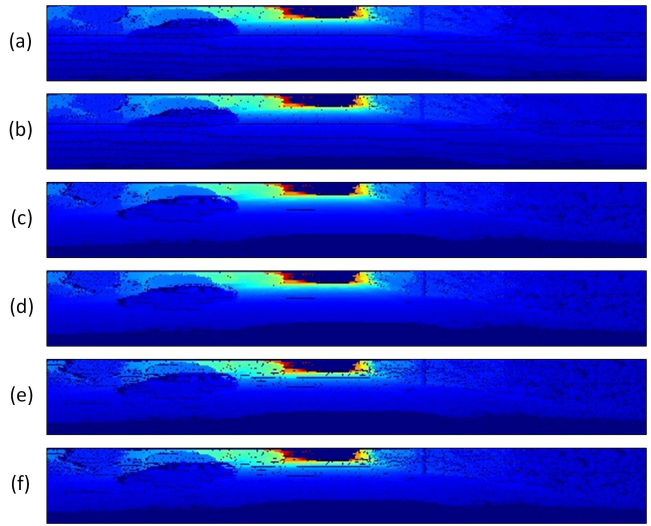}}
\end{center}
\caption{The range images in the current coordinate system. They are generated with different hardware (CPUs or GPUs) and LiDAR point clouds (current scan, previous scan, or previous range image). (a) and (b) are range images with current scan and based on CPUs and GPUs, respectively. (c) and (d) are range images with the previous scan and based on CPUs and GPUs, respectively. (e) and (f) are results with previous range images and based on CPUs and GPUs, respectively. }
\label{fig:rangedifference}
\end{figure}

Figure  \ref{fig:rangedifference} shows the range images generated with different source data and hardware. 
Among them, Fig. \ref{fig:rangedifference} (a) and (c) are the range images of the current and previous scans generated based on CPUs, respectively. The differences between the two range images are caused by the data type, the inaccurate relative pose, and the moving objects. The occlusion also results in more holes at the edge of the objects.
Figure \ref{fig:rangedifference} (c) and (e) are the results using the previous scan and previous range image on CPUs, respectively. We can find that due to the difference in point number, the range image generated by the previous range image is more sparse than the one generated by the previous scan.
Figure \ref{fig:rangedifference} (a) and (b) are the range images of the current scan based on CPUs and GPUs, respectively. The difference between the two range images is visually negligible, which indicates that the influence of the hardware on the range image generation process can be ignored.
Based on the above analysis and the qualitative comparison results in Fig. \ref{fig:rangedifference}, we finally use CPUs to convert the LiDAR scans into range images (like Fig. \ref{fig:rangedifference} (a)), and use GPUs to transform the features of the previous range images to the current range image (like Fig. \ref{fig:rangedifference} (f)).

\subsection{Evaluations on SemanticKITTI MOS Dataset}
In this section, we compare our method with the state-of-the-art MOS networks. Since there are few LiDAR-based MOS approaches, we also compare our method with some methods of semantic segmentation and scene flow.

Table \ref{tab:SOTA} shows the quantitative comparison results. \textit{SalsaNext(movable classes)} denotes the MOS result of directly marking all the movable classes as moving. \textit{SalsaNext(retrained)} retrains the SalsaNext network with binary MOS labels. \textit{SceneFlow}\cite{LiuQG19} uses the flow vector to determine the moving objects. \textit{SqSequence}\cite{ShiLWH020} and \textit{KPConv}\cite{ThomasQDMGG19} are multiple-scan-based LiDAR semantic segmentation networks. \textit{LMNet(N=1)}\cite{ChenLMWGBS21} uses range images and the precalculated residual image to differentiate the moving objects. \textit{LMNet(N=8 + Semantics)} represents the semantically enhanced version using 8 residual images. The proposed semantics-guided MOS network takes the semantic features of current range image, the transformed features of previous range image and the range images of current scan as inputs. 

As shown in Tab. \ref{tab:SOTA}, the performance of our semantics-guided MOS method is basically the same as \textit{KPConv}, slightly worse than the LMNet with 8 residual images and semantic information. 
The multiple-scan-based semantic segmentation method \textit{KPConv} is implemented based on point clouds, which requires high computational overhead and cannot achieve real-time performance. Our method only uses the compact LiDAR range images as inputs and can work in real-time.
The \textit{LMNet(N=8 + Semantics)} is equivalent to using 9 consecutive scans. Considerable computational overhead and time are spent on the generation of 8 residual images on CPUs. Our method only uses two adjacent LiDAR scans, and the feature association process (the ASA module) is implemented on GPUs.

\begin{figure*}[!t]
\begin{center}
\makebox[1pt]{\includegraphics[width=\linewidth]{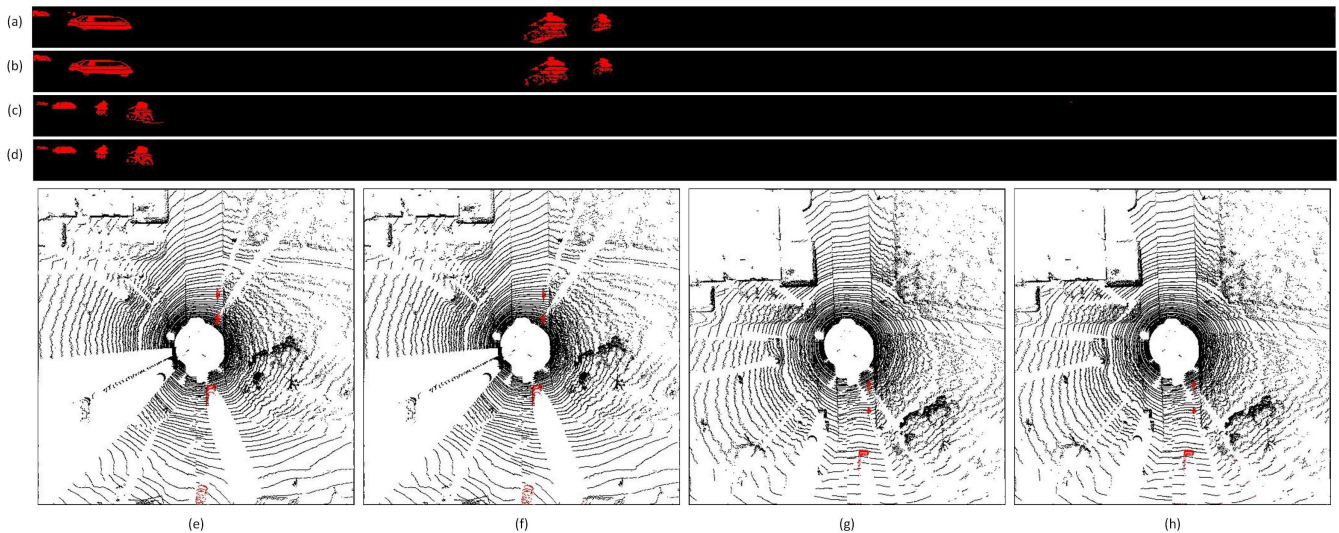}}
\end{center}
\caption{The MOS results in the forms of range image and point cloud. (b), (d), (f) and (h) are the MOS results predicted by our method. (a), (c), (e) and (g) are the corresponding ground truths. Red denotes the moving objects.}
\label{fig:mosresult}
\end{figure*}

Figure \ref{fig:mosresult} shows the ground truths and MOS results in the forms of range image and point cloud.

\begin{table}[!t]
\caption{MOS Performance Compared with the State-of-the-art (Test)}
\begin{center}  
\begin{tabular}{c c}
\hline
\textbf{Algorithms} & IoU\\
\hline
SalsaNext(movable classes) & 4.4\% \\
SceneFlow\cite{LiuQG19} & 4.8\% \\
SqSequence\cite{ShiLWH020} & 43.2\% \\
SalsaNext(retrained) & 46.6\% \\
KPConv\cite{ThomasQDMGG19} & 60.9\% \\
LMNet(N=1) & 52.0\% \\
LMNet(N=8 + Semantics) & 62.5\% \\
Ours & 60.6\% \\ 
\hline
\end{tabular}
\end{center}
\label{tab:SOTA}
\end{table}

\subsection{Ablation Study}
In order to save the computational overhead and time cost, we only use the training data sampled at equal intervals of 4000 scans in the ablation study section. The epoch size and batch size are set as 30 and 2, respectively.

\subsubsection{Effectiveness of Semantic Guidance}
Most of the existing LiDAR-based MOS methods use the raw LiDAR data and conduct the MOS task directly. The proposed network introduces a semantic segmentation module to assist the MOS process. In this section, we will analyze the effectiveness of the semantic guidance from the semantic segmentation module. 

Table \ref{tab:guidance} shows the MOS results on the validation dataset. \textit{MOS(RXYZI + Range Residual)} only uses the MOS module and takes range images and residual image as inputs. \textit{MOS(RXYZI + Range Residual) + Semantics} represents its semantically enhanced version and uses the semantic segmentation results at the end. \textit{MOS(RXYZI + Features)} denotes the proposed method with range images and semantic features as inputs of the MOS module.
Table \ref{tab:guidance} are the MOS results on the validation dataset with all training data. \textit{LMNet(N=1)} and \textit{LMNet(N=1 + Semantics)} use the precalculated residual image. The latter also uses the semantic segmentation results. \textit{MOS(RXYZI + Features)} represents the proposed network using range images and semantic features in the MOS module.

It is obvious that semantic information can help improve the MOS performance. The single-scan-based semantic segmentation task is closely related to the MOS task. Compared with using the residual images or using semantic segmentation results at the end, our method with semantic features as the inputs of the MOS module is more effective. The pre-trained single-scan-based semantic segmentation model can not only simplify the training process but also improve the MOS performance.

\begin{table}[!t]
\caption{Effectiveness of Semantic Guidance (Validation)}
\begin{center}  
\tabcolsep3pt   
\begin{tabular}{c c c}
\hline
\textbf{Algorithms} & Params & IoU\\
\hline
MOS(RXYZI + Range Residual) & 6711043 & 38.6\% \\
MOS(RXYZI + Range Residual) + Semantics & 13422615 & 41.4\% \\
MOS(RXYZI + Features) & 13423863 & 60.5\% \\
\hline
\end{tabular}
\end{center}
\label{tab:guidance}
\end{table}

\begin{table}[!t]
\caption{Effectiveness of Semantic Guidance (Validation)}
\begin{center}  
\tabcolsep3pt   
\begin{tabular}{c c c}
\hline
\textbf{Algorithms} & Params & IoU\\
\hline
LMNet(N=1) & 6711043 & 59.9\% \\
LMNet(N=1 + Semantics) & 13422615 & 61.4\% \\
MOS(RXYZI + Features) & 13423863 & 68.4\% \\
\hline
\end{tabular}
\end{center}
\label{tab:guidance2}
\end{table}

\subsubsection{Different MOS Inputs}
In this section, we will compare the MOS performance of different inputs, including raw LiDAR data (RXYZI), range residual image, and semantic features.

Table \ref{tab:inputs} shows the comparison results. \textit{RXYZI} gets the worst MOS performance due to the lack of temporal information. \textit{Features(current)} and \textit{RXYZI + Features(current)} also use the current LiDAR scan only, but obviously, semantic guidance without temporal information can still improve the results. \textit{RXYZI + Range Residual} and \textit{RXYZI + Features} combining the range images with the residual image or semantic features can increase the MOS accuracy. Compared with the residual image, the semantic features are more effective for the MOS task. However, using both residual image and semantic features at the same time degrades the performance. This may be caused by the conflicts between residual image and semantic features. A comparison between \textit{Features} and \textit{RXYZI + Features} shows that range images also contribute to the MOS task. Connecting the current semantic features with the transformed previous features can obtain better results than using the feature residuals or directly concatenating the semantic features without ASA operation. In summary, the transformed semantic features are more effective than the residual image and the original semantic features from previous scans. 

\begin{table}[!t]
\caption{Different MOS Inputs (Validation)}
\begin{center}  
\tabcolsep3pt   
\begin{tabular}{c c c}
\hline
\textbf{Algorithms} & Params & IoU\\
\hline
RXYZI & 6711011 & 26.9\% \\
Features(current) & 13423063 & 45.2\% \\
RXYZI + Features(current) & 13423223 & 51.8\% \\
\hline
RXYZI + Range Residual & 6711043 & 38.6\% \\
RXYZI + Features & 13423863 & 60.5\% \\
RXYZI + Range Residual + Features & 13423895 & 58.9\% \\
\hline
Features & 13423703 & 56.1\% \\
RXYZI + Feature Residuals & 13423223 & 39.1\% \\
RXYZI + Features(concat) & 13423863 & 48.7\% \\
\hline
\end{tabular}
\end{center}
\label{tab:inputs}
\end{table}

\subsubsection{Influence of Scan Numbers}
In this section, we will analyze the influence of the LiDAR scans used on the MOS performance. It is similar to the residual image analysis in LMNet. Theoretically, more LiDAR scans bring more useful information and improve the MOS performance.

Table \ref{tab:numbers} shows the MOS results with 2 to 8 consecutive LiDAR scans as inputs. It is obvious that the MOS accuracy does not increase with the increase of LiDAR scans. This is different from the residual images in LMNet. There may be two reasons. First, the semantic segmentation module is not accurate enough. The more LiDAR scans (semantic features), the more conflicts, which affects the improvement of the MOS result. Second, the relative poses between scans are also inaccurate. And this inaccuracy will increase with the increase of time interval.

In addition, although the transformations from the previous range images to the current are performed on GPUs, it still affects the training time cost. Based on the above considerations, we only use two adjacent LiDAR scans in our MOS method.

\begin{table}[!t]
\caption{Influence of Scan Numbers (Validation)}
\begin{center}  
\tabcolsep10pt   
\begin{tabular}{c c c}
\hline
\textbf{Scans} & Params & IoU\\
\hline
2 & 13423863 & 60.5\% \\
3 & 13424503 & 59.3\% \\
4 & 13425143 & 57.0\% \\
5 & 13425783 & 59.0\% \\
6 & 13426423 & 54.4\% \\
7 & 13427063 & 60.0\% \\
8 & 13427703 & 58.3\% \\
\hline
\end{tabular}
\end{center}
\label{tab:numbers}
\end{table}

\subsubsection{Influence of Modular Design}
The proposed network can be divided into the single-scan-based semantic segmentation module, the ASA module, and the MOS module. In order to analyze the benefits of modular design, we use different network architectures to replace the single-scan-based semantic segmentation module and the MOS module.

Table \ref{tab:modular} shows the MOS results with different network architectures. 
\textit{SS(RangeNet++)} and \textit{MOS(RangeNet++)} represent the use of RangeNet++ network in the single-scan-based semantic segmentation module and the MOS module, respectively. \textit{SS(SalsaNext)} and \textit{MOS(SalsaNext)} indicate the use of SalsaNext network. 

The SalsaNext network has better semantic segmentation performance than the RangeNet++ network. The MOS results in Tab. \ref{tab:modular} shows that due to the modular design, the MOS performance can be improved by simply updating any trainable module in the network.

\begin{table}[!t]
\caption{Influence of Modular Design (Validation)}
\begin{center}  
\begin{tabular}{c c c}
\hline
\textbf{Algorithms} & Params & IoU\\
\hline
SS(RangeNet++) + MOS(RangeNet++) & 100761335 & 29.5\% \\
SS(RangeNet++) + MOS(SalsaNext) & 57089655 & 48.4\% \\
SS(SalsaNext) + MOS(SalsaNext) & 13423863 & 60.5\% \\
SS(SalsaNext) + MOS(RangeNet++) & 57095543 & 35.7\% \\
\hline
\end{tabular}
\end{center}
\label{tab:modular}
\end{table}

\section{Conclusions}
In this paper, we propose a moving object segmentation network with semantic information as guidance. The whole network includes three modules: a single-scan-based semantic segmentation module, an adjacent scan association (ASA) module, and a multiple-scan-based moving object segmentation module. The ASA module works as an intermediate to connect the semantic segmentation module and the MOS module. After the ASA process, we can obtain the correspondences and the difference information between the semantic features, which are beneficial to the MOS task. The experiment on the SemanticKITTI MOS dataset shows the effectiveness of the network. The modular design also facilitates the upgrading of a single module and further improves the MOS accuracy.








\bibliographystyle{IEEEtran}
\bibliography{IEEEabrv,egbib}

\end{document}